\pdfoutput=1

\documentclass[11pt]{article}

\usepackage[final]{EMNLP2022} 

\usepackage{times}
\usepackage{latexsym}

\usepackage[T1]{fontenc}

\usepackage[utf8]{inputenc}

\usepackage{microtype}

\usepackage{inconsolata}

\usepackage{setspace}
\usepackage{graphicx}
\usepackage{makecell}
\usepackage{tabu, booktabs}
\usepackage{array,multirow}
\usepackage{tabularx}

\title{Toxicity Detection with Generative Prompt-based Inference}

\author{Yau-Shian Wang \and Yingshan Chang \\
  Carnegie Mellon University \\
  \texttt{\{yaushiaw, yingshac\}@cs.cmu.edu}}

\begin{document}
\maketitle
\begin{abstract}

\textit{Warning: this paper contains content that may
be upsetting or offensive.} \\
Due to the subtleness, implicity, and different possible interpretations perceived by different people, detecting undesirable content from text is a nuanced difficulty. It is a long-known risk that language models (LMs), once trained on corpus containing undesirable content, have the power to manifest biases and toxicity. However, recent studies imply that, as a remedy, LMs are also capable of identifying toxic content without additional fine-tuning. Prompt-methods have been shown to effectively harvest this surprising self-diagnosing capability. However, existing prompt-based methods usually specify an instruction to a language model in a discriminative way.
In this work, we explore the generative variant of zero-shot prompt-based toxicity detection with comprehensive trials on prompt engineering.
We evaluate on three datasets with toxicity labels annotated in social media posts. Our analysis highlights the strengths of our generative classification approach both quantitatively and qualitatively. Interesting aspects of self-diagnosis and its ethical implications are discussed.
\end{abstract}

\section{Introduction}

Language is notoriously powerful to project, spread and reinforce toxic opinions \cite{fiske1993controlling}. Since it is impossible to manually filter the massive online textual world, automatic toxicity detection is of utmost importance. 

\begin{figure}[t]
    \centering
    \includegraphics[width=\linewidth]{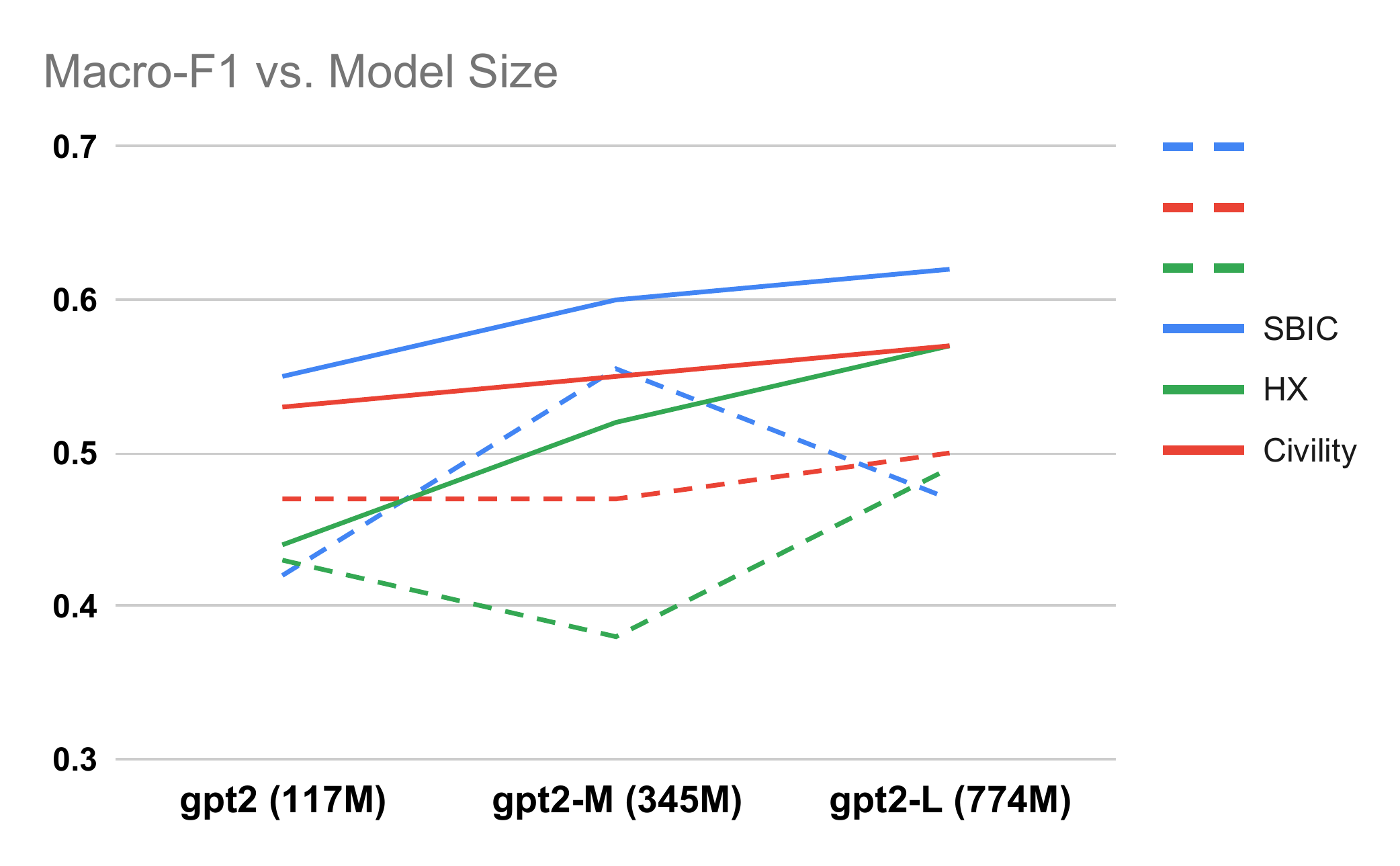}
    \caption{Macro-F1 achieved by GPT-2 with varying sizes on three datasets. Solid lines for \textit{Discriminative Classification} and dashed lines for \textit{Generative classification}. }
    \label{fig:model_sizes}
\end{figure}

Despite years of algorithmic advancement in language technologies, automatic toxicity detection still faces difficulties, the reason for which lies in three places. First, toxic language can be implicit. A harmful text can be harmless in terms of not containing profanity or slur words, but tends to spread hate and stereotypes by influencing people's judgments about others. This makes it hard both for data collection, since no explicit rules exist for crawling social media posts exhibiting implicit toxicity, and for modelling, because implicit toxicity is often beyond the reach of keyword-based methods \cite{prabhumoye2021few}. Second, mentions of minority groups often co-occur with toxicity labels scraped from online platform. Thus, it is hard to discourage a supervised model from exploiting the spurious correlation between mentions of minority groups and the toxicity label. Such undesirable exploitation, if went uncontrolled, could be damaging to members of minority groups. A third complicating factor is that subtle offensiveness might only be perceived by a certain group \cite{breitfeller2019finding}. This increases the detection difficulty because most detection models are trained to represent a generic perspective, which would down-weight the values of marginalized groups. 

On the high-level, this project aims to detect toxic language with zero-shot prompt-based inference. Recent works found that large language models (LLMs) acquire skills from their pre-training on automatically judging whether a text contains toxic content. This emerging behavior is presumably enabled by co-occurrences of \textit{<toxic language, textual judgment>} in the LLMs' pre-training corpus. We argue that guiding LLMs to detect toxicity in a zero-shot manner has the following merits: a)  Without supervision with binary labels, it is less likely that spurious feature-label associations can be injected into the model. b) Leveraging knowledge already acquired through pre-training reduces the effort for collecting data for task-specific supervision, which removes the burden in data-scarce scenarios. c) LLMs potentially demonstrate more robustness on long-tail phenomena where models relying too much on explicit keywords struggle to generalize. d) Relaxing the reliance on binary-label supervision is a key step to flexibly adapt a model to more nuanced labeling systems without re-training, which will prove crucial in developing responsible applications \cite{Toxigen}.

\section{Concrete Problem Definition}
\subsection{Task Formulation}
All tasks are formalized as text classification (binary) problems. Additionally, classification should be performed in a zero-shot manner without reliance on any toxicity-detection supervisions. Formally, given a text $x=\{w_1,w_2,\cdots,w_T\}$ with length $T$, we expect a binary label as an output indicating whether the input text contains toxic content.
For prompt-based methods (see Section~\ref{approach} for more details), in particular, we have a priori a specification of toxicity $y$, and obtain the classification label by prompting an LLM to answer a question about whether $x$ exhibits an aspect described by $d(y)$ \textbf{(discriminative classification)}, or prompting an LLM to generate $x$ on the condition of seeing a prefix $d(y)$ \textbf{(generative classification}).

\subsection{Datasets}
We consider three datasets containing social media posts and annotated toxicity labels: the \textsc{Social Bias Inference Corpus} (\textsc{SBIC}) \cite{sap2020social}, \textsc{HateXplain} (\textsc{HX}) \cite{HateXplain}, and \textsc{Civility} with offensiveness labels taken from the 2019 SemEval Task \cite{2019semeval} (the same dataset used in homework2).

\subsection{Metrics} We take \textsc{offensive} as the positive class. We report F1 for both positive and negative classes, as well as the macro F1. Our choice of evaluation metrics is based on two considerations. First, some models are not discriminative enough in a sense that they are predicting the same label all the time. If this label happens to be a majority label in a dataset, such an indiscriminative model can achieve an accuracy as high as 70\%, which is yet meaningless. In order to favor models whose performance is more balanced across two classes, we decided to abandon the accuracy measure and, instead, report F1 for the positive and negative classes separately. Second, the positive and negative classes in our datasets are not balanced. Therefore, a relative stronger performance on the dominating class could obscure the relative weakness on the other class. So, we argue it is more fair to use the macro-F1 as a single-number metric rather than the weighted average-F1.



\section{Related Work}

\paragraph{Toxicity detection datasets and methods}
Recent years have seen increased interest in hatespeech detection to combat the proliferation of toxic language spreading on social media platforms such as Twitter and Reddit. Many datasets have been annotated based on human-written social media posts \cite{founta2018large, waseem2016hateful}. Individual work might focus on one or several specific aspect(s) of toxicity. For instance, \citet{Kumar2018} annotated Facebook posts on covert aggression, \citet{Xu2012CyberBullying} studies cyber bullying.  \citet{zampieri-etal-2019-predicting} and \citet{ousidhoum-etal-2019-multilingual} provide multi-faceted labels regarding the targeted-nontargeted and individual-group distinctions. \citet{Georgakopoulos2018} explores a dataset where toxic comments are organized in six classes: toxic, severe toxic, obscene, threat, insult and identity attack. Various bias detection methods have been investigated in the literature, including keyword-matching \cite{Salminen2018AnatomyOO}, traditional machine learning classifiers (such as SVM \cite{davidson2017automated, malmasi2017detecting}, logistic regression \cite{Wulczyn2017Ex} and multi-layer perceptron \cite{Wulczyn2017Ex}), as well as Convolutional Neural Networks \cite{Georgakopoulos2018} and the most recent transformer-based approaches \cite{Mutanga2020HateSD, Glazkova2021FinetuningOP, Alonso2020HateSD}.

\paragraph{Prompt-based zero-shot inference}
Large language models (LLM) are found to master a sheer volume of implicit knowledge \cite{raffel2020exploring, wei2021finetuned}, making prompting a promising direction for addressing a wide spectrum of NLP tasks \cite{liu2021pre}. Prompting has successfully surfaced out hidden skills of LLMs in NLP sub-fields, such as sentiment classification \cite{seoh2021open}, question-answering \cite{yang2021empirical, dou2022zero}, social reasoning \cite{jiang2021delphi}, dialog generation \cite{zheng2021exploring, mi2021cins, madotto2021few} and image captioning \cite{tewel2021zero}. To draw a connection to computational ethics, \citet{schick2021selfdiagnosis} prompts GPT-2 and T5 for automated bias detection. \cite{prabhumoye2021few} extends this line of research using more structured prompts and performed few-shot experiments across different classes of LLMs with varying sizes. In this work, we extend the zero-shot toxicity detection approach explore in previous work \cite{schick2021selfdiagnosis} to its generative variant and demonstrate its greater competence.

\section{Baselines}

For a fair comparison with our zero-shot approach, we identify three approaches that also do not require task-specific supervision data as our baselines, namely ``toxicity lexicon", ``embedding similarity" and discriminative classification.

\paragraph{Toxicity Lexicon} We compile a toxicity lexicon from two sources \footnote{https://github.com/zacanger/profane-words/blob/master/words.json, https://www.freewebheaders.com/full-list-of-bad-words-banned-by-google/}. A text is classified as toxic if any word from the lexicon are is mentioned (case-insensitive).

\paragraph{Embedding Similarity} This baseline leverages a neural sentence encoder's \cite{gao2021simcse} ability to embed semantically relevant sentences closer together. In order to adhere to the requirement of not relying on any task-specific labels, we only compare the embedding similarities between a testing text $x$ and a description $d(y)$, where $y$ might claim that the text does or does not exhibit toxicity. 
\begin{equation}
    score_y = \mathit{sim}(E(x),E(d(y))),
\end{equation} 
$\mathit{sim}(\cdot)$ denotes cosine similarity and $E$ denotes the SinCSE sentence encoder.
The testing text is flagged as toxic if its embedding is closer to the ``toxic" description than the ``benign description", and vice versa.


\paragraph{Prompt-based Discriminative Classification}
In previous work \cite{schick2021selfdiagnosis}, text classification is re-formulated into a \textit{cloze test}, where an LLM $P_M$ is guided to predict an answer by completing a self-diagnosis prompt ($sdg(x,y)$). Such a prompt contains a testing text, $x$, concatenated with a binary question asking about whether an input text $x$ contains a type of toxicity $y$. 

\begin{quote}
    \colorbox{black!8}{x}\\
    \colorbox{black!8}{Question: Does the above text contain y?}\\
    \colorbox{black!8}{Answer: $\{$yes/no$\}$}
\end{quote}
Then, we re-interpret the output probabilities from the LLM into a probability of $x$ containing $y$:
\begin{equation} \label{eq:prompt}
    p(y|x) = \frac{p_M(Yes|sdg(x,y))}{\sum_{w \in \{Yes,No\}} P_M(w|sdg(x,y)) }.
\end{equation} 
Simply put, an LLM $P_M(\cdot)$ predicting ``Yes'' with a higher probability over predicting``No'' to the question implies that $x$ is more likely to contain $y$ (i.e. being toxic). 

\section{Approach}
\label{approach}
Recently, prompt-based learning has been widely applied for generating hatespeech or toxic language datasets~\cite{https://doi.org/10.48550/arxiv.2202.03286,https://doi.org/10.48550/arxiv.2203.09509}.
Their results imply that using some specific prompts can guide the model to generate harmful content.
Inspired by such a finding, we explore zero-shot prompt-based \textit{generative} classification with LLMs, which outputs the probability of generating a text conditioned on a prompt steering the LLM towards generating toxicity.
Compared with a non-toxicity-guiding prompt, if an toxicity-guiding prompt encourages an LLM to generate a testing text with higher probability, then this testing case is labeled as ``toxic''.

\begin{table*}[t!]
\small

\begin{tabularx}{\textwidth}{c|c|c|c|X}
\hline
ID & Model & Dataset & mac-F1 & \makecell{Prompt}\\
\hline
1 & Gen  & Civil & 0.48  & \makecell{$y^p$: Write a text that contains ethnic slur or toxic content, or insults other people\\
$y^n$: Write a text that doesn't contain ethnic slur nor toxic content nor insults other people\\}\\
\midrule
2 & Gen  & SBIC & 0.54 & \makecell{$y^p$: Write a text that contains ethnic slur or toxic content, or insults other people\\
$y^n$: Write a text that doesn't contain ethnic slur nor toxic content nor insults other people\\}\\
\midrule
3 & Gen  & HX & 0.54 & \makecell{$y^p$: Write a text that contains ethnic slur or insults black people\\
$y^n$: Write a text that doesn't contain ethnic slur nor toxic content nor insults black people\\}\\
\midrule
4 & Gen  & HX & 0.39 & \makecell{$y^p$: Write a text that contains offensive or toxic content\\
$y^n$: Write a text that doesn't contain offensive nor toxic content\\}\\
\midrule
5 & Gen  & HX & 0.43 & \makecell{$y^p$: Write a text that contains hatespeech\\
$y^n$: Write a text that doesn't contain hatespeech\\}\\
\midrule
6 & Gen  & Civil & 0.42  & \makecell{$y^p$: Write a text that contains offensive or toxic content\\
$y^n$: Write a text that doesn't contain offensive nor toxic content\\}\\
\midrule
7 & Dis  & SBIC & 0.47  & \makecell{Question: Does the above text contain very hateful, aggressive, disrespectful language?\\Answer:}\\
\midrule
8 & Dis & SBIC & 0.55 & Question: Does the above text exhibit intentional offensiveness or toxicity? Answer: \\
\midrule
9 & Dis & SBIC & 0.35 & \makecell{Question: Does the above text contain very rude, offensive, toxic, hateful, \\aggressive or disrespectful language?Answer:} \\
\bottomrule
\end{tabularx}
\caption{Prompts on different datasets and its perdformance. In the generative approach, we report the performance without providing task demonstration to the model.}
\label{tab:prompt}
\end{table*}

\subsection{Prompt-based Generative Classification}
Previous prompt-based text classification methods predicts a label $y$ in a \textit{discriminative} way that estimates $p(y|x)$.
We propose to classify a label in a \textit{generative} way~\cite{yogatama2017generative} that estimates the joint probability $p(x,y) \propto p(x|y)$.
Specifically, to classify $y$, we first convert $y$ into a pair of positive and negative prompts: 
\begin{quote}
    $y^{p}:$ Write a text that contains $y$.\\
    $y^{n}:$ Write a text that doesn't contain $y$
\end{quote}
, then the score estimating the log likelihood that an LLM generates $x$ conditioned on $y^p$ is:
\begin{equation} \label{eq:generative}
    s({y^{p}})=\sum_{t=1}^{T}\log p_M(x_t|y^{p},x_{<t}),
\end{equation}
then $p(y|x)=\frac{e^{s(y^p)}}{e^{s(y^p)}+e^{s(y^n)}}$.
Here, $p_M(\cdot)$ is a language model, such as GPT, which predicts next tokens auto-regressively.
If $x$ contains toxicity $y$, we expect $p_M(x|y^p) > p_M(x|y^n)$, i.e. $p(y|x)>0.5$. 

\subsection{Task Demonstration}
As indicated by using GPT-3 for few shot classification~\cite{NEURIPS2020_1763ea5a}, demonstrating a few examples sampled from the target task can improve the performance.
Here, we assume we have an unlabeled corpus $U$ for the target task.
We randomly sample a few texts as examples for guiding the model to generate texts similar to the texts in the target task.
Now, the score $s(y)$ becomes:
\begin{equation} \label{eq:demonstration}
    s({y^{p}})=\sum_{i=1}^{k} \sum_{t=1}^{T}\log p_M(x_t|y^{\cdot},x_{<t},u_i),
\end{equation}
where $k$ is the number of demonstrating examples and $u_i \sim U$ is an unlabeled text randomly sampled from $U$.
This idea is similar to model ensemble that we combine the scores conditioned on different examples.

\section{Results and Analysis}

\subsection{Prompt Engineering}
We find that both the discriminative and generative classification approaches are highly sensitive to the wording of the prompt.
In this section, we draw insight into what types of prompts boost performance.

\paragraph{Dataset Analysis} When designing a prompt, it is very important to understand the nature of the dataset.
We find that in SBIC dataset, any kind of offensive content will be labeled as ``offensive'', and there are wide range types of toxicity, such as sexual content, curse words, or racism.

On the other hand, in HX dataset, the types of hatespeech are monotonous that mainly consists of ethnic slurs, especially targeting at black people.
Only the texts contaiat ning discriminatory content are labeled as hatespeech.
For example, the sentence ``I fucking hate you. Are you retarded?'' is not labeled as hatespeech.
However the example ``You are a retarded nigger.'' is labeled as hatespeech because it involves racism.

In CIVILITY dataset, according to the statistic of their paper~\cite{2019semeval}, 1/3 toxic content targets at a group (hatespeech) and 2/3 toxic content targets at individual (cyberbulling).

\begin{table*}[th]
\small
\centering
\begin{tabular}{|l|ccc|ccc|ccc|}
\hline
 & \multicolumn{3}{c|}{SBIC} & \multicolumn{3}{c|}{HX} & \multicolumn{3}{c|}{Civility} \\
Approach & neg-F1 & pos-F1 & macro-F1 & neg-F1 & pos-F1 & macro-F1 & neg-F1 & pos-F1 & macro-F1 \\
\hline
Random & \multicolumn{3}{c|}{0.50} & \multicolumn{3}{c|}{0.50} & \multicolumn{3}{c|}{0.50} \\
Toxicity Lexicon & 0.58 & 0.52 & 0.55  & 0.47 & 0.51 & 0.49 & 0.82 & 0.48 & \textbf{0.65} \\ 
Embedding Similarity & 0.40 & 0.68 & 0.54  & 0.26  & 0.46 & 0.36 & 0.49 & 0.53 & 0.51\\
Discriminative Cls & 0.58 & 0.51 & 0.55 & 0.73 & 0.26 & 0.50 & 0.75 & 0.31 & 0.53 \\
\hline
Generative Cls & 0.71 & 0.36 & 0.54  & 0.74 & 0.33 & 0.54  & 0.77 & 0.21 & 0.48\\
Generative Cls+demo & 0.57 & 0.62 &  \textbf{0.60} & 0.69 & 0.44 & \textbf{0.57} & 0.64 & 0.50 & 0.57\\
\hline
\end{tabular}
\caption{Main results. For fair comparison,  we use gpt2-large for both discriminative and generative prompt-based classification models.}
\label{tab:compare_with_baseline}
\end{table*}

\paragraph{Prompt Design}
After understanding the nature of the dataset, we design a prompt for each dataset in Table~\ref{tab:prompt}.
The prompts with ID 1,2,3 are the prompts we use for each dataset.
For civil and SBIC, we use the same prompt because in both datasets the toxic content target at both individual and a group.
For HX, because it is a hatespeech dataset especially targeting at black people, we design a prompt that reflects this property.

By comparing ID 3 and ID 4 in Table~\ref{tab:prompt}, we find that including the term ``ethnic slur'' in our prompt greatly improves the performance because this term specifies the victims to be a group.
Using ``offensive or toxic content'' as prompt doesn't indicate the target of the victim, and thus leads to poor performance.
On the other hand, comparing ID 3 and ID 5, we find that LLMs cannot understand ``hatespeech'' is targeting a specific group, and thus also leads to poor performance.
Finally, the main difference between ID 1 and ID 6 is the term ``ethnic slur''. 
We find that adding this term encourages the model to detect the toxicity targeting a group, thus enhance the performance.

While \textit{generative classification} show promising results once a proper prompt is found, we experience a hard time making the \textit{discriminative classification} approach outperform random (F1$\sim$0.5). We posit that the primary advantage of \textit{generative classification} lies in ``improved contrastiveness", meaning that, as opposed to just predicting a Yes/No answer, the LLM is given more chances to distinguish between toxic vs. non-toxic with the prediction sequence is longer. ID 8 in Table~\ref{tab:prompt} is the best prompt we ever found that allows \textit{discriminative classification} to achieve a 0.55 on SBIC. Comparing ID 8 with ID 7, it seems that explicitly saying ``intend to offend" is useful in surfacing out the LLMs' underlying self-diagnosing abilities. Also interestingly, making a prompt wordy by listing adjectives describing what constitutes toxicity is counterproductive, resulting in the model more often predicting the opposite label in most of the times.
Comparing original \textit{generative} classification against \textit{generative classification + demo}, we find that it greatly improves positive F1, which implies it encourages the model to predict more accurate positive labels.
With a few demonstrating examples, the model learns how to generate texts that better match the distribution of the target task.

In conclusion, we find that the model is indeed very sensitive to the input prompt.
We need to design a prompt that specifies what we want from the model as precisely as possible, otherwise even a single ambiguous word could lead to in dramatic performance drop.

\subsection{Comparison Against Baselines}
Our main results against baselines are in Table~\ref{tab:compare_with_baseline}.
All of the unsupervised/zero-shot approaches only outperform the random baseline by a small margin, which means offensive content detection is still a very challenging task for LLMs.
Without finetuning, LLMs cannot understand the meaning of offensive content very well.
We speculate that this is because LLMs have less chance to touch toxic contents. 
When constructing the pre-training corpora for LLMs, researchers on purpose exclude these contents to prevent LLMs from learning some bad or harmful words. 

Our Toxicity Lexicon and Embedding Similarity baselines perform at a random level, with the exception that the Toxicity Lexicon achieves the highest score on the Civility dataset. This finding can be attributed to the fact that the data crawling process originally used to create the Civility dataset was guided by keyword matching.

Comparing our proposed generative approach against embedding similarity approach, our proposed method significantly improves the performance, especially on HX dataset.
We speculate the reason is that the sentence embedding without finetuning cannot capture the meaning of ``hatespeech''.

Our strongest baseline is the \textit{discriminative} prompt-based classification model.
In both \textit{discriminative} and our proposed generative models, for each dataset, we use the prompts that can achieve the best performance.
Our proposed method achieves better performance because \textit{generative} model allows more flexibility for LLMs to make a decision rather than a simple yes/no binary decision.
Conditioned on the offensive-guided prompt, any word, phrase or sentence that is relevant to the prompt naturally have higher probability to be generated.
Additionally, \textit{generative} classification is more similar to how we pre-train GPT-based models because both of them are in an autoregressive manner. 
LLMs may have not been trained on a yes/no question regarding toxicity or hatespeech.

We also analyze the impact of model size on the self-diagnosing ability. Figure~\ref{fig:model_sizes} reveals a positive model size-performance correlation for the \textit{generative} classification approach, but such a trend in \textit{discriminative} classification is unclear. We argue that the \textit{generative} classification results are noisier because those models mostly perform close to random, suggesting that when a prompt only asks a discriminative question, it fails to ``dig out" a model's potential in toxicity detection.

\subsection{Qualitative Analysis}

We use LIME \cite{LIME}, a model-agnostic interpretation approach to probe which tokens mainly influence an LLM's prediction. Table~\ref{tab:qualitative} show 4 examples where the prediction is correct and 8 examples where the prediction is wrong. The ``Bad Words" column shows words that exist in the toxicity lexicon. Colored highlights indicate tokens that greatly contribute to either a toxic (red) prediction or a non-toxic (blue) prediction, which is estimated by LIME. 

Example 1\&2 are correctly predicted with the correct toxic tokens identified. The model also did it right on Example 3\&4, even though slur words exists. This suggests that the existence of slur words does not correlate well with the toxicity label, confirming the need for a neural method with better context-awareness. 

On Example 5\&6 the model got it wrong because it put too much focus on benign words. Example 7\&8 demonstrates a reversed scenario, where the model made mistakes possibly because it got fooled by the existence of bad keywords. Example 9\&10 showcase a more challenging situation where hate is expressed highly implicitly, without mentioning any of the toxic keywords, which successfully confused the model. Finally, Example 11\&12 show non-toxic posts where toxic keywords are also absent, yet the model decided they were toxic. With the help of LIME, the model appears to think a certain hashtag and some non-toxic tokens somehow indicate toxicity, possibly because of confounding factors associating those benign tokens and toxicity labels together during the model's pre-training stage. We leave it for future work to further explore this intriguing misbehavior.

\newcommand{\blue}[1]{\textbf{\textcolor{blue}{#1}}}
\newcommand{\red}[1]{\textbf{\textcolor{red}{#1}}}

\begin{table*}[ht!]
\centering
\small
\begin{tabular}{|p{0.02\linewidth}|p{0.6\linewidth}|p{0.07\linewidth}|p{0.1\linewidth}|p{0.1\linewidth}|}
\hline

& Post & Label & Bad Words & Prediction \\
\hline
1 & i 'm a muslim man with a muslim wife . <\red{repeat}> . <\red{repeat}> he other day my wife called me a \red{paedophile},  so i said " that 's a big word for a five year old & 
T & paedophile & T \\
2 & RT @LivingBlueinRed: You know who never had to ask for immunity? \red{Hillary fucking Clinton} That's who. &
T & fucking & T \\
3 & This \blue{video} package is fucking me up.... dont do this to me... \#\blue{Wrestlemania} & N & fucking & N \\
4 & never underestimate the importance of a comfy office chair. especially if you're gonna be parking \blue{your} ass there for \blue{hours} & N & ass & N \\

\hline

5 & how \blue{many times} can you screw a blonde ? i do n't know , how much \blue{time} you got ? & T & screw & N \\
6 & Why do they put \blue{cotton} at the top of pill bottles? To remind blacks that they were \blue{cotten} pickers before they were pill pushers. & T & blacks & N \\
7 & I think I'm gonna cry again Another NYE alone I \red{hate} my life so much  Where's my New Year's Kiss and \red{party} Fuck & 
N & fuck & T \\
8 & A man already accused of killing two women and \red{suspected} in the deaths of as many as 30 more was charged with \red{raping} and murdering five other victims . & N & raping & T \\

9 & How do you turn a cat into a dragon? Tie it to the back of a car, it'll drag-on :') Heh I think I made that \blue{up}, tell me if you've heard it or something \blue{similar} :'\^) & T & & N \\
10 & I was hoping that at least one of those animals would get \blue{their} heads lobbed off by the propeller \blue{blades} like in Dawn of The Dead . & T & & N \\

11 & \#\red{HackAlert} Fri Mar 31 22:22:42 AEDT 2017 Failed \red{password} for root from 58.61.153.165 port 45977 ssh2 & N & & T \\
12 & \red{There} is a differance between waiting for you love and ignore the rest and waiting while you have fun with \blue{others}. & N & & T \\

\hline

\end{tabular}
\caption{Qualitative analysis on 1) when an LLM make mistakes, 2) whether the existence of toxic keywords interfere with LLM's decision, 3) which tokens have greater contribution to an LLM's decision-making (estimation made by LIME). T for toxic, N for non-toxic. Red tokens contribute to T-predictions and blue tokens contribute to N-predictions.}
\label{tab:qualitative}
\end{table*}

\section{Ethical Implications}
The intended usage of toxicity detection is to aid the identification of undesirable content in social media or other text-based landscapes. Our approach is able to detect implicit toxic contents not captured by slur keywords with better data-efficiency. Additionally, zero-shot approaches are less biased against minority groups because they can be more easily steered towards a specific toxic aspect via flexible prompt design without risking exploiting spurious MinorityGroupMention-and-label associations.

Despite the benevolent aspect of this work, we are also aware of its potential negative impacts and we welcome voices from all sides to join the discussion. To begin with, the success of self-diagnosis relies on the assumption that language models have been trained on co-occurrences of unjust content with associated textual judgments (e.g. "This sentence contains content that may be offensive or toxic."). However, the increasingly careful examination and filtering of pre-training corpus could remove such co-occurrences. While we do not deny the necessity of corpus filtering for social good, the point we want to make is that data-filtering may come at a cost. It is unclear whether in the future, when the pre-training corpus are cleaner and absent of toxic content, the self-diagnosing ability as a pre-training by-product will be wiped off. More studies have to be done on the understanding of the dynamics between what are fed into the language models and what types of awareness or capabilities emerge as results.

Moreover, an LLM's ability to self-identify toxicity stems from its own ability to generate toxicity. This could lead to a dual-use issue, where  prompt-based methods can be misused intentionally or unintentionally to introduce offensive language to targeted groups. 

Additionally, due to the lack of comprehensive interpretation of how LLMs leverage its own black-box knowledge for self-diagnosis, there could be potential risks of biases inside a bias detection model \cite{sap2019risk}. For example, media posts generated by a certain demographic group could induce more false positive flags versus posts written by another group. Thus, we invite future work to inspect where the implicit knowledge of "What constitutes toxicity" in LLMs originated from in the first place, as a crucial step to transparent model behaviors in bias detection.

\section{Conclusion}
Automatic toxicity detection facilitates online moderation, which is an important venue for NLP research to positively impact the society. Recent works \cite{prabhumoye2021few, schick2021selfdiagnosis} demonstrate that large-scale pre-trained language models are able to detect toxic contents without fine-tuning. Prompts can be carefully designed to harness the implicit knowledge about harmful text learned by MLM pre-training. Such a zero-shot detection manner not only effectively avoids the model from picking up spurious feature-label correlations introduced by additional supervision signals, but also enables flexible instruction-giving on the fly. In this project we expand the investigation on zero-shot \& prompt-based toxicity detection from discriminative classification to generative classification. We incorporate the \textsc{Social Bias Frame} \cite{sap2020social} into our prompt design, facilitating the detection of implicit biases. We evaluation our approaches on three hatespeech/toxicity detection datasets. Our quantitative and qualitative results demonstrate the advantage of generative classification over discriminative classification. Finally, we provide analysis on the interaction between the wording of prompts and the behavior of models, followed by discussions on the ethical implications of self-diagnosis.

\bibliography{custom}
\bibliographystyle{acl_natbib}


\end{document}